%% file: main.tex
\documentclass[letterpaper, 10 pt, conference]{ieeeconf}

\IEEEoverridecommandlockouts
\usepackage{microtype}
\usepackage{graphicx}
\usepackage{booktabs} 
\usepackage{amsmath,amsthm,amsfonts}
\usepackage{bm}
\usepackage{color}
\usepackage{algorithm}
\usepackage{algorithmic}

\usepackage{hyperref}

\newtheorem{theorem}{Theorem}
\newcommand{\name}[0]{Proximal Deterministic Policy Gradient}
\newcommand{\nameac}[0]{{PDPG}}

\title{\LARGE \bf
Proximal Deterministic Policy Gradient
}

\author{Marco Maggipinto $^{1}$, Gian Antonio Susto$^{1}$, and Pratik Chaudhari$^{2}$
\thanks{*This work was not supported by any organization}
\thanks{$^{1}$Marco Maggipinto and Gian Antonio Susto are with the Department of Information Engineering, University of Padova, Padova, Italy
         \href{mailto:marco.maggipinto@dei.unipd.it}{\tt\small marco.maggipinto@dei.unipd.id}
         \href{mailto:gianantonio.susto@dei.unipd.it}{\tt\small gianantonio.susto@dei.unipd.it}}%
\thanks{$^{2}$Pratik Chaudhari is with the Electrical and Systems Engineering Department, University of Pennsylvania, Philadelphia, PA, USA
         \href{mailto:pratikac@seas.upenn.edu}{\tt\small pratikac@seas.upenn.edu}}%
}

\begin{document}

\maketitle
\thispagestyle{empty}
\pagestyle{empty}

\begin{abstract}

This paper introduces two simple techniques to improve off-policy Reinforcement Learning (RL) algorithms. First, we formulate off-policy RL as a stochastic proximal point iteration. The target network plays the role of the variable of optimization and the value network computes the proximal operator. Second, we exploits the two value functions commonly employed in state-of-the-art off-policy algorithms to provide an improved action value estimate through bootstrapping with limited increase of computational resources. Further, we demonstrate significant performance improvement over state-of-the-art algorithms on standard continuous-control RL benchmarks.

\end{abstract}
\section{Introduction} \label{intro}
\input{sections/intro.tex}

\section{Related work} \label{related}
\input{sections/related.tex}

\section{Background} \label{back}
\input{sections/back.tex}

\section{Proximal gradient methods} \label{proximal}
\input{sections/proximal.tex}

\section{Proposed Method} \label{method}
\input{sections/proposed.tex}

\section{Experimental results} \label{res}
\input{sections/experimental.tex}

\section{Ablation study} \label{ablres}
\input{sections/ablation.tex}

\section{Conclusions} \label{concl}
\input{sections/concl}

\addtolength{\textheight}{-8cm}   

\section*{APPENDIX}
\subsection{Training details}
We employed Feedforward Neural Networks with two hidden layers of 256 neurons each with ReLu \cite{glorot2011deep} activations for both policy and critic.

As in \cite{fujimoto2018addressing} we perform a burn-in at the beginning of training where we sample random actions from the environment. The hyper-parameters employed are listed in Tables \ref{tab:fixed_hyper}, \ref{tab:var_hyper}. The exploration noise, action noise and noise clip are relative to the maximum value of the action that varies depending on the environment.
\begin{table}[!h]
\centering
\caption{Hyper-parameter values used in all the environments.}
\label{tab:fixed_hyper}
\begin{tabular}{lr}
\toprule
\textbf{Name}          & \textbf{Value} \\
\midrule
Learning rate $\alpha$                  & 3e-4  \\    
Damping coefficient $\tau$                    & 0.005 \\
Exploration noise $\sigma$                  & 0.1   \\
Noise clip $c$ & 0.5 \\
Action noise $\sigma_q$ & 0.2   \\
Batch size                & 256   \\
Proximal steps $n_{prox}$                & 5  \\
Policy loss strength $\beta$                & 0.01  \\
\bottomrule
\end{tabular}
\end{table}

\begin{table}[!h]
\centering
\caption{Additional Hyper-parameter values.}
\label{tab:var_hyper}
\begin{tabular}{llr}
\toprule
\textbf{Environment} & \textbf{Name}          & \textbf{Value} \\
\midrule
Hopper, Walker & Burn-in                  & 1000 \\
All others & Burn-in                & 10000 \\ \hline
HalfCheetah, Ant & Policy Weight Decay                  & 0.0 \\
All others & Policy Weight Decay                  & 1e-5 \\ \hline
Humanoid & Proximal Strength ($1/\lambda)$                  & 10.0 \\
Hopper, Walker & Proximal Strength ($1/\lambda)$                  & 1.0 \\
HalfCheetah, Ant & Proximal Strength ($1/\lambda)$                  & 0.1 \\
\bottomrule
\end{tabular}
\end{table}


\bibliographystyle{IEEEtran}
\bibliography{main.bib}
\end{document}

%% file: sections/intro.tex

Actor Critic (AC) \cite{Konda2000} algorithms have become the de facto standard in continuous control Reinforcement Learning tasks allowing the employment of powerful function approximation methods such as Deep Neural Networks (DNNs) to directly learn the control policy. While very effective in simulated environments, poor sample efficiency have limited their deployment on real systems, where querying the environment for new samples is expensive.

High variance of the gradient estimate \cite{sutton2018reinforcement} is at the foundation of such inefficiency. Algorithms like TRPO \cite{schulman2015trust} and PPO \cite{schulman2017proximal} operate in a sample regime that fail to provide good approximations of the true gradient \cite{ilyas2018deep} with considerable impact on performance, moreover, their on policy nature requires new data to be collected at each optimization step wasting all past transitions.
Deterministic Policy Gradient (DPG) algorithms \cite{silver2014deterministic} improve upon these methods by employing deterministic policies and off-policy updates; the former limit the source of randomness to the sole environment with a consequent reduction in the number of samples required for gradient estimation and the latter allows for data reuse by storing past transitions in a replay buffer and employing them during the entire training procedure.

Another source of error in policy updates is a poor action value function estimate. Here multiple factors come into play. On the one hand, Overestimation Bias \cite{thrun1993issues} causes Q-learning algorithms to exhibit a consistent overestimation of the action value, with potentially divergent errors; to mitigate such problem, Double Q-learning \cite{van2016deep} employs two independent Q-functions trained with a mixed update, however,  \cite{fujimoto2018addressing} showed that this approach is not suitable in an AC setting and proposes a similar solution where the Time Difference (TD) update is performed with the minimum of the two action value functions.
\\ On the other hand, the bootstrapping nature of TD updates results in a regression problem that changes over time making the optimization procedure very tricky and even unstable (if combined with off-policy updates and function approximation in the infamous deadly triad \cite{sutton2018reinforcement}). Most AC algorithms employ target networks that are slowly updated during training to provide a stable regression target. This approach was initially introduced in Double Deep Q-Networks (DDQN) \cite{van2016deep} and since then it has been adopted by Deep Deterministic Policy Gradient (DDPG) \cite{lillicrap2015continuous}, Twin Delayed Deep Deterministic policy
gradient (TD3) \cite{fujimoto2018addressing}, Soft Actor Critic (SAC) \cite{haarnoja2018soft}. Target networks play a fundamental role in the optimization procedure and algorithms are typically very sensible to the speed which the networks are updated at.

In a Deep RL setting, TD learning is performed by minimizing a surrogate loss function (typically the Mean Square Error) with Stochastic Gradient Descent (SGD) based algorithms, the most common choice is Adam \cite{kingma2014adam} that has proven effective for training DNNs. In this work, we propose an alternative optimization procedure to tackle the sample efficiency problem that provides a principled interpretation of target networks and minimizes a single loss function combining both policy and value updates. Our procedure employs \textit{Time Damped} Stochastic Proximal Gradient (SPG) \cite{parikh2014proximal} \cite{ryu2014stochastic} iteration, widespread in convex optimization, combined with bootstrapped action value estimates and is able to provide improved performance compared to state-of-the-art algorithms on continuous control tasks.
More in details, we endow TD3 with a proximal gradient optimization procedure and we exploit the two Q-networks already used in the original algorithm to limit overestimation bias also to provide a more accurate action value estimate via bootstrapping that allows better policy updates.

%% file: sections/related.tex

The first successful application of DNNs in RL dates back to \cite{mnih2013playing} where Deep Q-learning was introduced to Play Atari games at human level capabilities. Target networks where first introduced in \cite{van2016deep} where they proposed an improved version of Deep Q-learning for the same control task. Since then, such algorithm has been the reference for Q-function estimation with DNNs.

In continuous control tasks, Q-learning methods are not enough to learn a policy; in fact, finding the maximizing action would require the solution of a maximization problem every time the agent needs to act on the environment, with prohibitive computational costs. Here, AC algorithms comes into play where a parametrized policy learns to maximize the total expected reward. These methods typically follows the policy iteration \cite{sutton2018reinforcement} paradigm where at each time step the Q-function is estimated and then the policy is made greedy w.r.t. it. Hence Deep Q-learning is still a fundamental part of methods such as DDPG \cite{lillicrap2015continuous}, A3C \cite{mnih2016asynchronous}, TD3 \cite{fujimoto2018addressing}, SAC \cite{haarnoja2018soft} that are all related to our method. Alternative approaches that follows the AC paradigm but employ sample estimates of the Q-function can be found in \cite{schulman2015high}, TRPO \cite{schulman2015trust}, PPO \cite{schulman2017proximal} and P3O \cite{fakoor2019p3o}.

There has been attempts to improve AC algorithms from an optimization point of view, by proposing alternatives to the TD error with more complex loss functions; SBEED \cite{dai2017sbeed} provides a primal dual interpretation of the Bellman Equation that results in a minmax game to optimize the convex dual of the quadratic loss function. \cite{feng2019kernel} proposes a kernel loss alternative to the MSE that enables improved training of the Q-function.
\\ Proximal methods, while widespread in convex optimization, have been used to train DNNs only in \cite{chaudhari2018deep} in a Supervised Learning setting.

%% file: sections/back.tex

RL deals with the problem of learning a maximally rewarding behavior for an agent interacting with its environment. Formally, this can be cast in the framework of optimal policy estimation in a Markov Decision Process (MDP) \cite{bertsekas2019reinforcement}. A MDP is a tuple $(S, A, p(\bm{s}'|\bm{s}), r(\bm{s}, \bm{a}), \gamma)$ where $S$ is the set of states that the environment can assume, $A$ the set of actions that can be performed on the environment, $p(\bm{s}'|\bm{s}, \bm{a})$ the probability of transitioning to state $\bm{s}'$ after taking action $\bm{a}$ in state $\bm{s}$, $r(\bm{s}, \bm{a})$ the reward function, that can be either deterministic or stochastic, and $\gamma$ the discount factor used to weight future rewards and guarantee finite total rewards even for infinite time horizon problems.
The goal of an RL algorithm is finding a policy $\pi(\bm{a} | \bm{s})$ that maximizes the total expected discounted reward:
\begin{equation} \label{eq:reward}
    J = E_{\pi}\left[\sum_{t=1}^{T} \gamma^t \, r(\bm{s}_t, \bm{a}_t)\right]
\end{equation}
Where the expectation is taken over all sources of randomness in the MDP.

Policy Gradient methods tackle the problem by directly maximizing (\ref{eq:reward}) with respect to the parameters $\bm{\phi}$ of a NN parametrizing the policy function, using gradient based optimization procedures. We know from the policy gradient theorem \cite{sutton2000policy} that it is possible to express the gradient of (\ref{eq:reward}) with respect to the policy parameters as an expectation over trajectories:
\begin{equation}\label{eq:policygrad}
    \nabla J(\bm{\phi}) = \mathbb{E}_{\pi} \left[\sum_{t=1}^{T} Q^{\pi}(\bm{s}_t, \bm{a}_t) \log \pi_{\bm{\phi}}(\bm{a}_t | \bm{s}_t)\right]
\end{equation}

Where the Q-function (or critic) $Q(\bm{s}, \bm{a})$ is the total expected reward obtained starting from state $\bm{s}$, performing action $\bm{a}$ and then following policy $\pi$. The Q-function can be estimated using TD learning, exploiting the Bellman equation:
\begin{equation} \label{eq:ballman}
    Q^{\pi}(\bm{s}, \bm{a}) = r + \gamma \mathbb{E}_{\pi}\left[Q^{\pi}(\bm{s'}, \bm{a'})\right]
\end{equation}
Here $r$ is the expected reward after taking action $\bm{a}$ in state $\bm{s}$. To simplify the optimization procedure and speed-up policy updates, the expectation over trajectories in (\ref{eq:reward}) is typically replaced with an expectation over transitions, hence maximizing the marginal expected reward. The resulting policy gradient is as follows:
\begin{equation}
        \nabla J(\bm{\phi}) = \mathbb{E}_{\pi} \left[Q^{\pi}(\bm{s}, \bm{a}) \log \pi_{\bm{\phi}}(\bm{a}| \bm{s})\right]
\end{equation}
In on policy methods such as PPO the actions are sampled from the current policy while off-policy methods employs a replay buffer $\mathcal{B}$ where past transitions are stored.

The policy gradient theorem stated in (\ref{eq:policygrad}) is valid for stochastic policies; there is an analogous for the deterministic case \cite{silver2014deterministic} where the policy gradient can simply be obtained by backpropagation through the Q-function; then gradient ascent steps are taken in order to make the policy greedy with respect to the actual estimate of the action values. When NNs are used as function approximators for the Q-functions, the Ballman update in (\ref{eq:ballman}) requires itself the minimization of a loss function called TD error with respect to the network parameters $\bm{\theta}$. The resulting algorithm solves the coupled optimization problem:
\begin{equation}
\begin{aligned}
        \bm{\theta}^* = & \underset{\bm{\theta}}{\textrm{argmin}} \, E_{(\bm{s}, \bm{a}, \bm{s}')\sim \mathcal{B}} \left[TD_{\bm{\theta}}(\bm{s}, \bm{a}, \bm{s}') \right] \\
      \bm{\phi}^* = & \underset{\bm{\phi}}{\textrm{argmax}} \, E_{(\bm{s}, \bm{a}) \sim \mathcal{B}} \left[Q_{\bm{\theta}}^{\pi}(\bm{s}, \pi_{\bm{\phi}}(\bm{a}|\bm{s}) \right]
\end{aligned}
\end{equation}
With $TD_{\bm{\theta}}(\bm{s}, \bm{a}, \bm{s}') = || Q_{\bm{\theta}}(\bm{s},\bm{a}) - r - Q_{\bm{\theta}}(\bm{s}', \pi_{\bm{\phi}}(\bm{a}|\bm{s}')) ||^2$.
Typically, to make SGD steps more stable, target networks are used both for the policy and the action value with parameters $\bm{\phi}'$ and $\bm{\theta}'$ that are averaged exponentially over time: $\bm{\phi}' = \tau \bm{\phi} + (1 - \tau)\bm{\phi}'$,  $\bm{\theta}' = \tau \bm{\theta} + (1 - \tau) \bm{\theta}'$ with $\tau << 1$. The TD error is then computed as $TD_{\bm{\theta}}(\bm{s}, \bm{a}, \bm{s}') = || Q_{\bm{\theta}}(\bm{s},\bm{a}) - r - Q_{\bm{\theta}'}(\bm{s}', \pi_{\bm{\phi}'}(\bm{a}|\bm{s}')) ||^2$ with gradient propagation only on $\bm{\theta}$.

%% file: sections/proximal.tex

SGD based optimization algorithms perform exceptionally well at training DNNs especially in a Supervised setting where the data distribution does not change during training. In RL, however, this assumption is not true since data are collected with different policies at different time instants and the TD error targets evolve during time; all these factors make the optimization procedure difficult. Proximal Methods have shown appealing convergence and stability properties in convex optimization and can be an alternative to standard gradient based algorithms. Given a function $f: \mathbb{R}^N \rightarrow \mathbb{R}$ and a constant $\lambda \in \mathbb{R}$ we define the proximal operator for a point $\bm{y} \in \mathbb{R}^N$ as:
\begin{equation} \label{eq:proximal}
    prox_{\lambda f}(\bm{y}) = \underset{\bm{x}}{argmin} f(\bm{x})  + \frac{1}{2\lambda} || \bm{x} - \bm{y} ||^2
\end{equation}
For a convex function $f$ the following theorem holds:
\begin{theorem}
A vector $\bm{x}^* \in \mathbb{R}^N$ is a critical point for the function $f$ iff $\bm{x}^* = prox_{\lambda f}(\bm{x}*)$
\end{theorem}
For a detailed proof we refer the reader to \cite{parikh2014proximal}. This implies that, starting from a random point $\bm{x}_0$, by repeated applications of the proximal operator one can hope to reach the minimum of $f$; for this to happen, $prox_{\lambda f}(\bm{x})$ must be a contraction. While this is not true, the map is a firm non-expansion i.e. for every $\bm{x}, \bm{y} \in \mathbb{R}^N$:
\begin{equation}
    ||\Delta||^2 \leq (\bm{x} - \bm{y})^T\Delta
\end{equation}
With $\Delta = prox_{\lambda f}(\bm{x}) - prox_{\lambda f}(\bm{y})$. \\
Defining the damped proximal iteration with constant $\tau$ as:
\begin{equation}
    \bm{x}_{k+1} = \tau \bm{x}_{k} + (1 - \tau) prox_{\lambda f}(\bm{\bm{x}_{k}})
\end{equation}
This iteration provably converges to a stationary point hence, in the convex setting, proximal iteration is an effective optimization method with convergence guarantees.

In the stochastic non-convex setting, the statements above do not hold true but the algorithm has still some desirable properties that make it a valid alternative to SGD. More in details, being $f_k$ the loss function associated to the batch sampled at time $k$, the Stochastic Proximal Iteration (SPI), starting from a random point $\bm{x}_0$ is:
\begin{equation}
    \bm{x}_{k+1} = prox_{\lambda f_{k}}(\bm{x}_k)
\end{equation}
SPI can be interpreted as SGD performed on a smoothed loss function  derived from the viscosity solution of the Hamilton Jacobi equation \cite{chaudhari2018deep} $u(\bm{x},t)$  defined as:
\begin{equation}
    u(\bm{x},t) = \underset{\bm{y}}{min} f(\bm{y}) + \frac{1}{2t}||\bm{x} - \bm{y}||^2
\end{equation}
In fact, if $prox_{tf}(\bm{x})$ exists, it is true that:
\begin{equation}
    \nabla u(\bm{x}, t) = \frac{\bm{x} - prox_{tf}(\bm{x})}{t}
\end{equation}
Hence, performing SGD updates on $u(\bm{x}, t)$ results in a \textit{damped} SPI on the function $f$:
\begin{equation} \label{eq:dampedprox}
    \begin{split}
        \bm{x}_{k+1} = & \, \bm{x}_k - \nabla u_k(\bm{x}, t) \\
        =& \, \bm{x}_{k} -  \frac{\bm{x_k} - prox_{tf_k}(\bm{x_k})}{t} \\
        = & \, (1- \frac{1}{t}) \bm{x}_k + \frac{1}{t} prox_{tf_k}(\bm{x_k})
    \end{split}
\end{equation}
Here $t$ plays the role of $\lambda$ in (\ref{eq:proximal}) and also serves as exponential averaging constant of the \textit{damped} SPI. In particular, for $t=1$ we recover SPI with $\lambda=t=1$. As shown in \cite{chaudhari2018deep} the function $u(\bm{x}, t)$ has smoother local minima and it's easier to optimize. For this reason, SPI has better stability and convergence properties than SGD.

%% file: sections/proposed.tex
In this section we provide a detailed description of our algorithm and its optimization procedure. The proposed method is based on TD3 that has proven effective at solving continuous control tasks and provide state-of-the-art performance while being simple and easily reproducible. We provide here an interpretation of target networks that play a fundamental role in the optimization procedure of AC algorithms and are a widespread trick to make training more stable. Such interpretation can be easily derived from a few simple changes to (\ref{eq:dampedprox}); given variables $\bm{x}$ and $\bm{x'}$, we rewrite (\ref{eq:dampedprox}) as:

\begin{equation} \label{eq:proxsys}
\begin{cases} \bm{x}_{k+1} = prox_{\lambda f_{k}}(\bm{x}'_{k}) \\ \bm{x}'_{k+1} = \tau \bm{x}_{k+1} + (1 - \tau) \bm{x}'_{k} \end{cases}
\end{equation}
Where we have introduced two hyper-parameters, $\lambda$ and $\tau$ that control respectively the proximal term strength and the damping constant. This decouples the two terms as opposed to (\ref{eq:dampedprox}) where the single parameter $t$ controls both, giving more freedom to tune the algorithm behavior.
It is immediately clear how the time evolution of $\bm{x}$ in (\ref{eq:proxsys}) is analogous to the parameters evolution during training of the "fast" moving function. Similarly $\bm{x}'$ plays the role of the target parameters that slowly change during time. The hyper-parameter $\lambda$ allows to control how close the two remains, which may help trading-off the update speed and how off-policy the data collected are.
As in TD3 we employ target functions for the policy and the two action value networks with parameters denoted respectively as $\bm{\phi}'$, $\bm{\theta}_1'$ and $\bm{\theta}_2'$. The pair of Q-functions is used to reduce the overestimation bias and also to provide a more reliable estimate of the action value by bootstrapping; to train the two Q-networks we minimize the following TD error: 
\begin{equation}
    TD_{\bm{\theta}_i}(\bm{s}, \bm{a}, \bm{s}') = huber(Q_{\bm{\theta}_i}^{\pi}(\bm{s}, \bm{a}) - y)
\end{equation}
with $y=r + \underset{i \in {1,2}}{min} \, Q_{\bm{\theta}'_i}(s', \pi_{\bm{\phi}'}(\bm{s}))$. \\ We employ the smooth-$L1$ loss (or huber) instead of the MSE. This choice is justified by the nature of the Bellman equation (\ref{eq:ballman}): the expected value over the next state and action is estimated in the TD error with a single transition and thus present a high variance; the huber loss put less weight on large errors compared to the MSE trusting less the expectation estimate. Moreover, since the targets change during training, the smooth-$L1$ loss may improve stability reducing strong changes in the parameters during a single optimization step.
The policy network is trained to maximize the average action value of the two target Q-functions; the corresponding loss for a single transition $\ell_{\bm{\phi}}(\bm{s})$ is:
\begin{equation}
    \ell_{\bm{\phi}}(\bm{s}) = - 0.5\left(Q_{\bm{\theta}'_1}^{\pi}(\bm{s},  \pi_{\bm{\phi}}(\bm{s})) + Q_{\bm{\theta}'_2}^{\pi}(\bm{s}, \ \pi_{\bm{\phi}}(\bm{s})\right)
\end{equation}
There are two main differences in the loss functions compared to standard TD3: \begin{enumerate}
    \item Target networks are used to compute the policy gradients instead of their "fast" counterpart.
    \item A bootstrapped estimate of the action value leverages both the the available Q-networks to reduce the approximation error.
\end{enumerate}
The first difference implies that our method does not require delayed policy updates because the target networks change slowly during time. Moreover, the improved quality of the action value gives better gradient estimates for the policy. The resulting methods thus performs SPI on a single loss function $\ell(\bm{\theta}, \bm{\phi})$:
\begin{equation} \label{eq:loss}
\begin{split}
        \ell(\bm{\theta}_1, \bm{\theta}_2, \bm{\phi}) = E_{(\bm{s}, \bm{a}, \bm{s}')\sim \mathcal{B}}  \left[ \right. &TD_{\bm{\theta}_1}(\bm{s}, \bm{a}, \bm{s}')  \\  + & TD_{\bm{\theta}_2}(\bm{s}, \bm{a}, \bm{s}') \\ + &  \beta\, \ell_{\bm{\phi}}(\bm{s}) \left. \right] 
\end{split}
\end{equation}
Where we introduced the hyper-parameter $\beta$ to control the scale of the two loss functions. We believe this expedient is important since the TD error has the same scale of the one step reward while the Q-function that is maximized by the policy has magnitude similar to the cumulative reward. For most Mujoco environments the difference is of almost three orders of magnitude.

As in \cite{fujimoto2018addressing} we add clipped noise to the actions in order to smooth the action value functions. In particular, for each action $\bm{\bar{a}}$ in the batch $B_k$ we sample noise from a normal distribution $\bm{\epsilon} \sim \mathcal{N}(\bm{0}, \sigma_q \bm{I})$ and then set:
\begin{equation} \label{eq:clip}
    \bm{\bar{a}} = \bm{\bar{a}} + clip(\bm{\epsilon}, -c, c)
\end{equation} 
Here $c$ is an hyper-parameter and the clipping is performed element-wise on the vector $\bm{\epsilon}$.

The SPI procedure detailed in (\ref{eq:proxsys}) requires for each batch the computation of the proximal operator. This can be done through full gradient descent on the loss function defined by the batch sampled from the replay buffer at time step $k$. More in details, we run $n_{prox}$ gradient descent steps to minimize the proximal loss defined as:
\begin{equation}
\begin{split}
    \mathcal{L}^{prox}_k = \ell_{k}(\bm{\theta}_1, \bm{\theta}_2, \bm{\phi})  + \frac{1}{2\lambda}\left( \right. & ||\bm{\theta}_1 - \bm{\theta}'_{1, k}||^2 \\ +   &||\bm{\theta}_2 - \bm{\theta}'_{2, k}||^2 \\
        +  &||\bm{\phi} + \bm{\phi}'_k||^2 \left. \right)
\end{split}
\end{equation}
We use a single hyper-parameter $\lambda$ to control the strength of the proximal term for both the policy and the Q-networks. In our implementation, we replace the $L_2$ norm with the MSE in order to have all the proximal terms scaled with respect to number of parameters.
 \\
The resulting algorithm, called \name{} (\nameac),  is summarized in Algorithm \ref{alg:td3prox}. 
\begin{algorithm}[tb]
   \caption{\nameac}
   \label{alg:td3prox}
\begin{algorithmic}
   \STATE {\bfseries Input:} $\tau$, $n_{prox}$, $\lambda$,  batch
   size $n_B$, learning rate $\alpha$, exploration noise variance $\sigma$
     \STATE Initialize network parameters $\bm{\theta}_1$, $\bm{\theta}_2$, $\bm{\phi}$ randomly
   \STATE $\bm{\theta}'_1 \gets \bm{\theta}_1$, $\bm{\theta}'_2 \gets \bm{\theta}_2$, $\bm{\phi}' \gets \bm{\phi}$
   \REPEAT
      \FOR{$k=1$ {\bfseries to} $T$}
        \STATE Collect transition $(\bm{s}, \bm{a}, \bm{s}', r)$ with exploratory action $\bm{a}=\pi_{\bm{\theta}}(\bm{s}) + \epsilon$ with $\epsilon \sim \mathcal{N}(0, \sigma\bm{I})$ 
        \STATE Store transition in the replay buffer $\mathcal{B}$
        \STATE
        \STATE Sample minibatch of $n_B$ transitions from the replay buffer $B_{k} \sim \mathcal{B}$
        \STATE Add clipped noise to actions in minibatch as in  (\ref{eq:clip})
        \FOR{$i=1$ {\bfseries to} $n_{prox}$}
            \STATE $\Delta \bm{\theta}_1$, $\Delta \bm{\theta}_2$, $\Delta \bm{\phi}$ = $\nabla \mathcal{L}^{prox}_k$
            \STATE $\bm{\theta}_1 \gets \bm{\theta}_1 - \alpha\Delta\bm{\theta}_1$
            \STATE $\bm{\theta}_2 \gets \bm{\theta}_1 - \alpha\Delta\bm{\theta}_2$
            \STATE $\bm{\phi} \gets \bm{\phi} - \alpha\Delta\bm{\phi}$
        \ENDFOR
        \STATE $\bm{\theta}'_1 \gets \tau \bm{\theta}_1 + (1-\tau)\bm{\theta}'_1$
        \STATE $\bm{\theta}'_2 \gets \tau \bm{\theta}_2 + (1-\tau)\bm{\theta}'_2$ 
        \STATE $\bm{\phi}' \gets \tau \bm{\phi} (1-\tau) \bm{\phi}'$
       \ENDFOR
\UNTIL{convergence}
\end{algorithmic}
\end{algorithm}

%% file: sections/experimental.tex

\begin{table*}[!h]
\renewcommand{\arraystretch}{1.4}
\centering
\caption{Average time-steps required by each algorithms to reach the reward thresholds set approximately at one third and two thirds of the maximum reward achieved by the best algorithm.}
\label{tab:efficiency}
\begin{tabular}{lrrrrrrrrrr}
\toprule
{} & \multicolumn{2}{c}{\textbf{Ant}} & \multicolumn{2}{c}{\textbf{HalfCheetah}} & \multicolumn{2}{c}{\textbf{Hopper}} & \multicolumn{2}{c}{\textbf{Humanoid}} & \multicolumn{2}{c}{\textbf{Walker2d}} \\
Thresholds &  2000  &   4000  &       5000  &  10000 &  1000  &  2000  &    3000  &   6000  &    1800  &  3600  \\
\midrule
Ours & 371500 &  \textbf{743500} &      \textbf{101000} &  \textbf{490500} &  \textbf{92500} & \textbf{133000} &   390000 & \textbf{1426000} &   \textbf{134500} & \textbf{227000} \\
SAC  & \textbf{326000} & 1135000 &      122000 &  624000 & 203000 & 299000 &   \textbf{288000} & 2335000 &   314000 & 540000 \\
TD3  & 432000 & 2002000 &      147500 & 1392500 & 147000 & 202000 &   402500 & 1750000 &   187500 & 365000 \\
PPO  & 977306 &  822067 &         / &     / & 190874 & 359629 &      / &     / &   484557 & 231834 \\
\bottomrule
\end{tabular}
\end{table*}
\begin{figure*} [!h]
    \centering
    \includegraphics[width=1.0\textwidth]{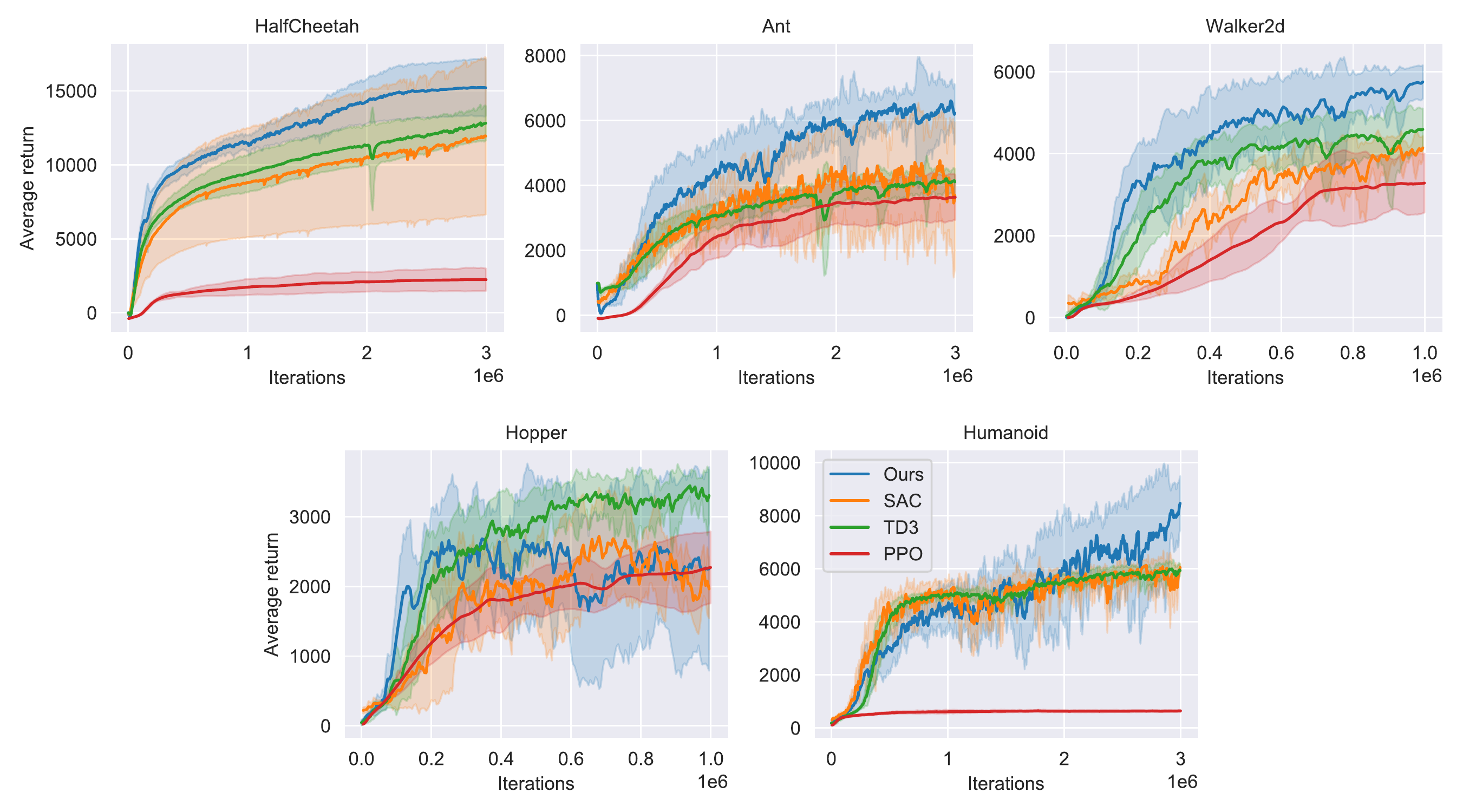}
    \caption{Training curves on OpenAI gym continuous control benchmarks. Our methods consistently outperform concurrent approaches (SAC and TD3) and on policy methods (PPO).}
    \label{fig:results}
\end{figure*}

\begin{figure*} [!h]
    \centering
    \includegraphics[width=1.0\textwidth]{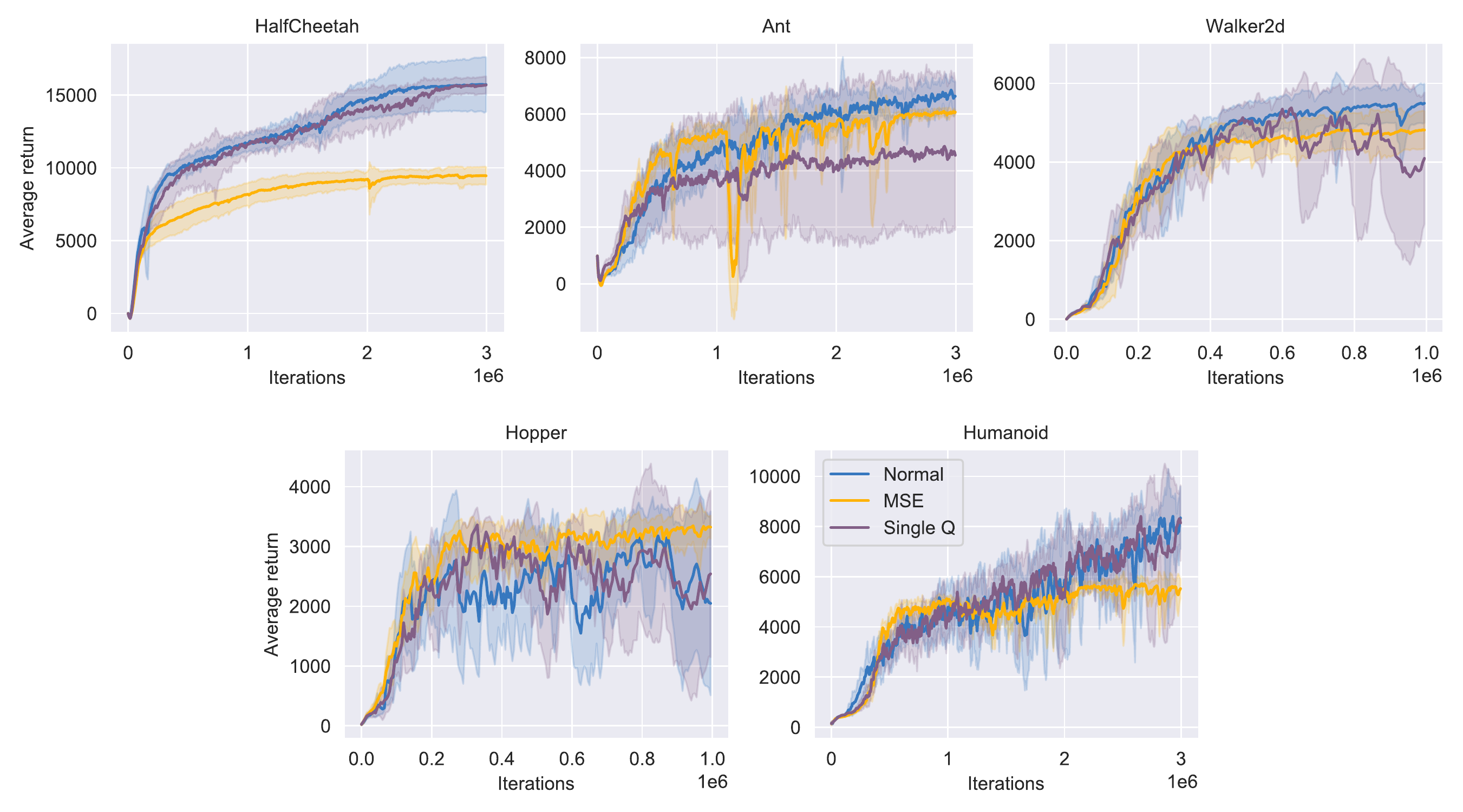}
    \caption{Ablation study comparing the training curves of our method (blue) and its versions using the MSE in the TD error (yellow) and no bootstrapped value estimate (purple).}
    \label{fig:resultsabl}
\end{figure*}

We provide in this section a comprehensive performance analysis of the proposed method to assess its capabilities in terms of sample efficiency and asymptotic performance with particular focus on the former being of fundamental interest in real applications where querying the environment for new samples is extremely expensive. In such a scenario being able to trade-off sample requirements with policy optimality is fundamental; in fact, it may be preferable to reach reasonable performance with few samples than having high asymptotic capabilities.

We train our agent on Mujoco \cite{todorov2012mujoco} OpenAI gym \cite{brockman2016openai} continuous control tasks, a challenging benchmark often used in literature to test RL algorithms dealing with continuous action and state spaces. We compare our method against state-of-the-art on-policy (PPO) and off-policy (TD3, SAC) algorithms. PPO \cite{schulman2017proximal} is an on-policy algorithm that exploit the policy gradient theorem and generalized advantage estimation \cite{schulman2015high} to learn an optimal policy. During training, a trust region constraint is imposed on the policy that is updated keeping it close to the one at the previous iteration; this results in more stable training and better performance. TD3 \cite{fujimoto2018addressing} is the algorithm which our method is based on, it learns a deterministic policy exploiting the deterministic policy gradient theorem and additional tricks describes in Section \ref{method} to improve upon DDPG \cite{lillicrap2015continuous} which we do not include in our comparison being very similar to TD3 with lower performance. SAC \cite{haarnoja2018soft} is a maximum entropy RL algorithm that learns a stochastic policy to maximize the total expected reward plus an entropy term in order to achieve high performance while being maximally exploring. TD3 and SAC are the most performing algorithms and show similar results.

Figure \ref{fig:results} shows a comparison of the training curves of the algorithms listed above. We run our algorithm for ten different seeds in order to assess its stability with different conditions; the reward is averaged among ten episodes, keeping the default maximum episode length defined by the gym framework. In each plot, the solid lines represent the average value among different seeds while the shaded area indicates a single standard deviation from the average. The curves have been smoothed for visual clarity with an exponential average. \\ For TD3 we have reported the curves taken from the authors github repository \footnote{https://github.com/sfujim/TD3} except for the environments where the authors provided curves for only 1 million steps (Ant, HalfCheetah) or didn't provide them at all (Humanoid); for such cases we run the experiments with the code available in the same repository. For SAC we employed the curves available at the project website \footnote{
https://sites.google.com/view/soft-actor-critic} and used the OpenAi baselines \footnote{https://github.com/openai/baselines} code to run experiments with PPO. \\ It is noticeable how our method consistently outperform both on policy and off-policy methods on most continuous control tasks. The Hopper environment exhibits a quite noisy behavior but the average performance are comparable with the other algorithms. Moreover, it takes much less for our method to reach an average return equal to the maximum performance obtained by the other methods. See for example the HalfCheetah environment where \nameac{} is able to match the performance of TD3 with half the number of samples.

To better characterize sample efficiency we report in Table \ref{tab:efficiency} the average number of time-steps required by each algorithm to exceed a set of reward thresholds placed at approximately one third and two thirds of the maximum reward achieved by the best algorithm. The superior sample efficiency of our method here is evident, for most of the environments \nameac{} reaches the specified thresholds with much less samples than the others. In the Humanoid and Ant it shows less efficiency than SAC for the lower threshold but better efficiency for the higher one, moreover, it has consistently better asymptotic performance.
We acknowledge that this sample efficiency comes at a cost: the computation of the proximal operator requires multiple gradient steps for each batch, slowing down the training; in our experiments we took five gradient steps for each batch hence the amount of computations required scaled accordingly. We believe that this is a fair price to pay since it reduces significantly the number of queries to the environment needed to train the agent properly.

%% file: sections/ablation.tex
We provide in this section an ablation study to show the effect on the proposed method of the huber loss and the bootstrapped value estimation. We run a version of the method that employs the MSE in the TD error and also a version that doesn't use the bootstrapped estimation, but still keeps two Q-functions to avoid the overestimation bias. 

In Figure \ref{fig:resultsabl} the training curves are reported for the two alternative versions compared to the standard approach. The employment of bootstrapped value estimates while not drastically changing performance seems to provide improved stability, this can be seen especially from the Ant and Walker environment where the Single Q algorithm has much higher variance. 
The huber loss has a drastic effect on the HalfCheetah environment, providing considerable performance improvement. On the remaining environments there is not substantial difference between the two loss functions. 

In general, the standard method shows better performance and stability hence all the components employed are empirically justified by this study.

%% file: sections/concl.tex

In this paper we proposed \name{}, an off-policy RL method for model free continuous control tasks that exploits proximal gradient methods and bootstrapping to better solve the TD error optimization problem. Proximal algorithms are appealing in an RL setting since they show improved convergence and stability properties compared to standard SGD. Moreover, we showed that proximal methods provide a natural interpretation of the target networks, a trick commonly employed in RL to stabilize training. \\ The resulting algorithm compare favourably with state-of-the-art off-policy and on-policy methods showing improved sample efficiency and asymptotic performance. The significant increase in sample efficiency makes our algorithm appealing for deployment in real environments, this possibility will be explored in a future work.